\documentclass[times, 10pt,twocolumn]{article} 
\usepackage{latex8}

\usepackage{amsmath}
\usepackage{amssymb,bm}
\usepackage{graphicx}
\usepackage{subfig}

\usepackage{tikz}
\usetikzlibrary{arrows,automata}
\usepackage{multirow}
\usepackage{xspace}  

\usepackage{wrapfig}
\usepackage{todonotes}

\usepackage[]{algorithm2e}

\usepackage{mathtools}
\DeclarePairedDelimiter\ceil{\lceil}{\rceil}

\usepackage{perpage}
\MakePerPage{footnote}

\newcommand{\ie}{\textit{i.e.}\xspace}

\newcommand{\eg}{\textit{e.g.}\xspace}

\newcommand{\pos}[2]{\textit{pos}_{#1,#2}}
\newcommand{\posb}[3]{\textit{pos}_{#1,#2,#3}}
\newcommand{\dpart}[2]{d_{#1,#2}}
\newcommand{\dtot}[1]{d_{#1}}
\newcommand{\dist}{D}
\newcommand{\nm}[2]{n_{#1,#2}}
\newcommand{\m}[1]{m_{#1}}
\newcommand{\M}{Z}

\usepackage[latin1]{inputenc}

\begin{document}
\pgfdeclarelayer{background}
\pgfdeclarelayer{foreground}
\pgfsetlayers{background,main,foreground}

\title{On the Synthesis of Guaranteed-Quality Plans for Robot Fleets in \\Logistics Scenarios via Optimization Modulo Theories}

\author{%
	Francesco Leofante\\
	 Erika {\'{A}}brah{\'{a}}m\\
	RWTH Aachen University\\
	\and 
	Tim Niemueller\\
	Gerhard Lakemeyer\vphantom{\'{A}}\\
	RWTH Aachen University\\
	\and
	Armando Tacchella\\
	University of Genoa
}

\maketitle
\thispagestyle{empty}

\begin{abstract}
   In manufacturing, the increasing involvement of autonomous robots in production processes poses new challenges on the production management. In this paper we report on the usage of Optimization Modulo Theories (OMT) to solve certain multi-robot scheduling problems in this area. Whereas currently existing methods are heuristic, our approach guarantees optimality for the computed solution. We do not only present our final method but also its chronological development, and draw some general observations for the development of OMT-based approaches.
\end{abstract}

\section{Introduction}
\label{sec:intro}

With the advent of Industry 4.0, factories are moving from static process chains towards more autonomy, by introducing robots in their production lines. Given this paradigm shift, the problem of managing and optimizing the in-factory supply chain carried out by (fleets of) autonomous robots becomes crucial.

The RoboCup Logistics League (RCLL)~\cite{PlanningInRCLL} has been proposed as a realistic testbed to study the above mentioned problem at a comprehensible and manageable scale. There, groups of robots need to maintain and optimize the material flow according to dynamic orders in a simplified factory environment.

Though there exist successful symbolic reasoning methods towards solving the underlying scheduling problem \cite{DBLP:conf/aaai/HofmannNCL16,DBLP:conf/aaaiss/NiemuellerLF13}, a disadvantage of these methods is that they provide no guarantees about the quality of the solution. A promising solution to this problem is offered by the recently emerging field of Optimization Modulo Theories (OMT), where Satisfiability Modulo Theories (SMT) solving is extended with functionalities towards optimization~\cite{DBLP:conf/sat/NieuwenhuisO06,DBLP:journals/tocl/SebastianiT15,DBLP:conf/tacas/SebastianiT15}.

In this paper we report on the application of OMT to compute optimal strategies for multi-robot systems within the RCLL scope. We encode the underlying scheduling problem as a linear mixed-integer problem which can be solved by OMT solvers such as \texttt{Z3}~\cite{Z3}, \texttt{SMT-RAT}~\cite{DBLP:conf/sat/CorziliusKJSA15} and \texttt{OptiMathSAT}~\cite{DBLP:conf/cav/SebastianiT15}. 

We integrated our OMT-based planning module in the RCLL planning framework~\cite{SMTInRCLL}. By rigorous experimental analysis we show that naive encodings fail to cope with the complexity of the domain. We then detail our findings and solutions adopted to overcome previous limitations. 

After presenting some preliminaries in Section~\ref{sec:preliminaries}, we specify the problem we are going to solve in Section~\ref{sec:robots}. We explain our OMT-based solution and provide experimental evaluations in Section~\ref{sec:encoding}. Finally, we draw some general conclusions and discuss future directions of research in the Sections \ref{sec:optim} and \ref{sec:conclusion}.
\section{Preliminaries}
\label{sec:preliminaries}

\paragraph{Mixed-integer arithmetic.} Problems considered in this work are encoded as \emph{mixed-integer arithmetic} formulas.
Syntactically, (arithmetic) \emph{terms} are constant symbols,
variables, and sums, differences or pro\-ducts of terms.  (Arithmetic)
\emph{constraints} compare two arith\-metic terms using $<$, $\leq$,
$=$, $\geq$ or $>$. (Quantifier-free arithmetic) \emph{formulas} use
conjunction $\wedge$ and negation $\neg$ (and further syntactic sugar
like disjunction $\vee$ or implication $\rightarrow$) to combine
theory constraints. Formulas that do not use multiplication are called
\emph{linear}. A formula in \emph{conjunctive normal form}
(\emph{CNF}) is a conjunction of disjunctions of theory constraints or
negated theory constraints (see Eq. (\ref{eq:abs}) for a simple example
formula in CNF).  Semantically, each variable is interpreted over
either the real or the integer domain by an \emph{assignment},
assigning to each variable a value from its domain; we use the
standard semantics to evaluate formulas.

\paragraph{Optimization modulo theories.}

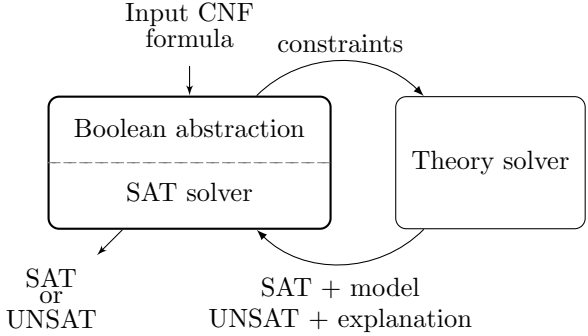
\begin{figure}[t]
	\begin{center}
		\scalebox{1}{\begin{tikzpicture}[thick]


\tikzstyle{doc}=[%
draw,
thick,
align=center,
color=black,
shape=document,
minimum width=10mm,
minimum height=15.2mm,
shape=document,
inner sep=2ex,
]

    \node [rectangle, draw,text centered, rounded corners, text width=10em, minimum height=5em, minimum width = 10em] (sat) {    	
    	\begin{tabular}{c}
    	Boolean abstraction\\
    	\\
    	SAT solver
    	\end{tabular}};

    \node[above of=sat, node distance=1.8cm](cnf){\begin{tabular}{c}
    	Input CNF\\[-0.08cm]
    	formula
    	\end{tabular}};
    
    \begin{pgfonlayer}{foreground}
    \path (sat.west |- sat.west) node (a) {};
    \path (sat.east -| sat.east)node (b) {};
    \path[ draw=black!50, dashed]
    (a) rectangle (b);
    \end{pgfonlayer}
    
    \node [rectangle, draw,text centered, rounded corners, text width=6.5em, minimum height=5em, minimum width =6.5em, right of=sat, node distance=4cm] (smt) {Theory solver};

	\node[below left of=sat, node distance=2.6cm](res){\begin{tabular}{c}
		SAT \\[-0.2cm]
		or\\[-0.1cm]
		UNSAT
		\end{tabular}};

	\path [draw, -latex'] (cnf) -- (sat);
	\draw[-latex,bend left=45]  (sat) edge [above] node {constraints} (smt);
	\draw[-latex,bend right=-45]  (smt) edge  [below] node {\begin{tabular}{c}
		SAT + model \\
		UNSAT + explanation
		\end{tabular}} (sat);
	\path [draw, -latex'] (sat) -- (res);
	
\end{tikzpicture}}
	\end{center}
	\vspace*{-2ex}
	\caption{The SMT solving framework.}
	\label{fig:smt}
\end{figure}

%
\emph{Satisfiability modulo theories} (\emph{SMT}) \emph{solving}
aims at deciding the
satisfiability of (usually quantifier-free) first-order logic formulas over
some theories like,
\eg, 
the theories of lists, arrays, bit vectors, real or (mixed-)integer arithmetic.  
To decide the satisfiability of an input formula $\varphi$ in CNF, SMT solvers (see Fig. \ref{fig:smt}) typically first build a \emph{Boolean abstraction} $\textit{abs}(\varphi)$ of $\varphi$ by replacing each constraint by a fresh Boolean variable (proposition), \eg,
\vspace*{-1ex}
\begin{eqnarray}
\arraycolsep=2pt
\begin{array}{ccccccccccc}
\varphi &= &x \geq y &\wedge &(&y > 0& \vee &x >0&)& \wedge &y \leq 0 \\
\textit{abs}(\varphi)&=&A& \wedge& (&B& \vee &C&)& \wedge  &\neg B 
\end{array}
\label{eq:abs}
\end{eqnarray}
where $x$ and $y$ are real-valued variables, and $A$, $B$ and $C$ are propositions.

\noindent A \emph{Boolean
satisfiability} (\emph{SAT}) \emph{solver} searches for a
satisfying assignment $S$ for $\textit{abs}(\varphi)$, \eg, $S(A)=1$, $S(B)=0$,
$S(C)=1$ for the above example.  If no such assignment exists then the
input formula $\varphi$ is unsatisfiable. Otherwise, the
consistency of the assignment in the underlying theory
is checked by a \emph{theory solver}. In our example, we check whether the set $\{ x \geq y,\ y
\leq 0,\ x > 0\}$ of linear inequalities is feasible, which is the
case. If the constraints are consistent then a satisfying solution
(\textit{model}) is found for $\varphi$. Otherwise, the theory solver returns a theory lemma $\varphi_E$ giving an
\textit{explanation} for the conflict, \eg, the negated conjunction some inconsistent input constraints.
The explanation is used to refine the Boolean abstraction $\textit{abs}(\varphi)$ to $\textit{abs}(\varphi)\wedge \textit{abs}(\varphi_E)$.
These steps are iteratively executed until either a theory-consistent
Boolean assignment is found, or no more Boolean satisfying assignments
exist. \emph{Less lazy} variants invoke theory checks more frequently also on partial assignments.

\emph{Optimization modulo theories} (\emph{OMT})
\cite{DBLP:conf/tacas/BjornerPF15,DBLP:conf/sat/CorziliusKJSA15,DBLP:conf/cav/SebastianiT15,DBLP:conf/tacas/SebastianiT15}
extends SMT solving with optimization procedures to find a variable assignment that 
defines an optimal (say minimal) value for an objective function $f$ (or a 
linear, lexicographic, Pareto, etc. combination of multiple
objective functions) under all models of a formula $\varphi$. Most OMT solvers implement the 
\textit{linear-search} scheme \cite{DBLP:journals/tocl/SebastianiT15}, which first uses SMT solving to determine a model $S$ for $\varphi$.
Let $\varphi_S$ be the conjunction of all theory constraints that
are true under $S$ and the negation of those that are false under $S$. In minimization problems, a minimum $\mu$ for $f$ is computed under the
side condition $\varphi_S$, and $\varphi$ is updated to 
\vspace*{-1ex}
\begin{equation}
\label{eq:optimization}
\varphi \wedge (f < \mu) \wedge \neg \varphi_{S}
\end{equation}

\noindent Repeating this procedure until the formula becomes unsatisfiable will lead to an assignment minimizing $f$ under all models of $\varphi$.

\paragraph{Planning with SMT solving.}
SMT solvers are nowadays embedded as core engines
in a wide range of technologies (see \eg
\cite{DBLP:conf/sefm/AbrahamK16} for some examples).
%
In the area of planning, \cite{DBLP:conf/icra/NedunuriPMCK14} and
\cite{DBLP:conf/aips/WangDCK16} use SMT solving to generate task and
motion plans.
The authors of \cite{DBLP:conf/rss/DantamKCK16} perform task and motion planning which leverages
incremental solving in \texttt{Z3} to update constraints about motion
feasibility. The work \cite{DBLP:conf/iros/SahaRKPS14} presents a motion
planning framework where SMT solving is used to combine motion
primitives so that they satisfy some linear temporal logic (LTL) requirements.
 
In contrast to the above works, $(i)$ we do not use additional knowledge (\eg, motion
primitives, plan outlines) to seed the search performed by the SMT
solver and $(ii)$ we exploit OMT solving to synthesize plans that are not only \textit{feasible} but
also \textit{optimal}.

\section{Logistics Robots in Simulation}
\label{sec:robots}

\begin{figure}[t]
	\begin{center}
		\includegraphics[width=0.45\textwidth]{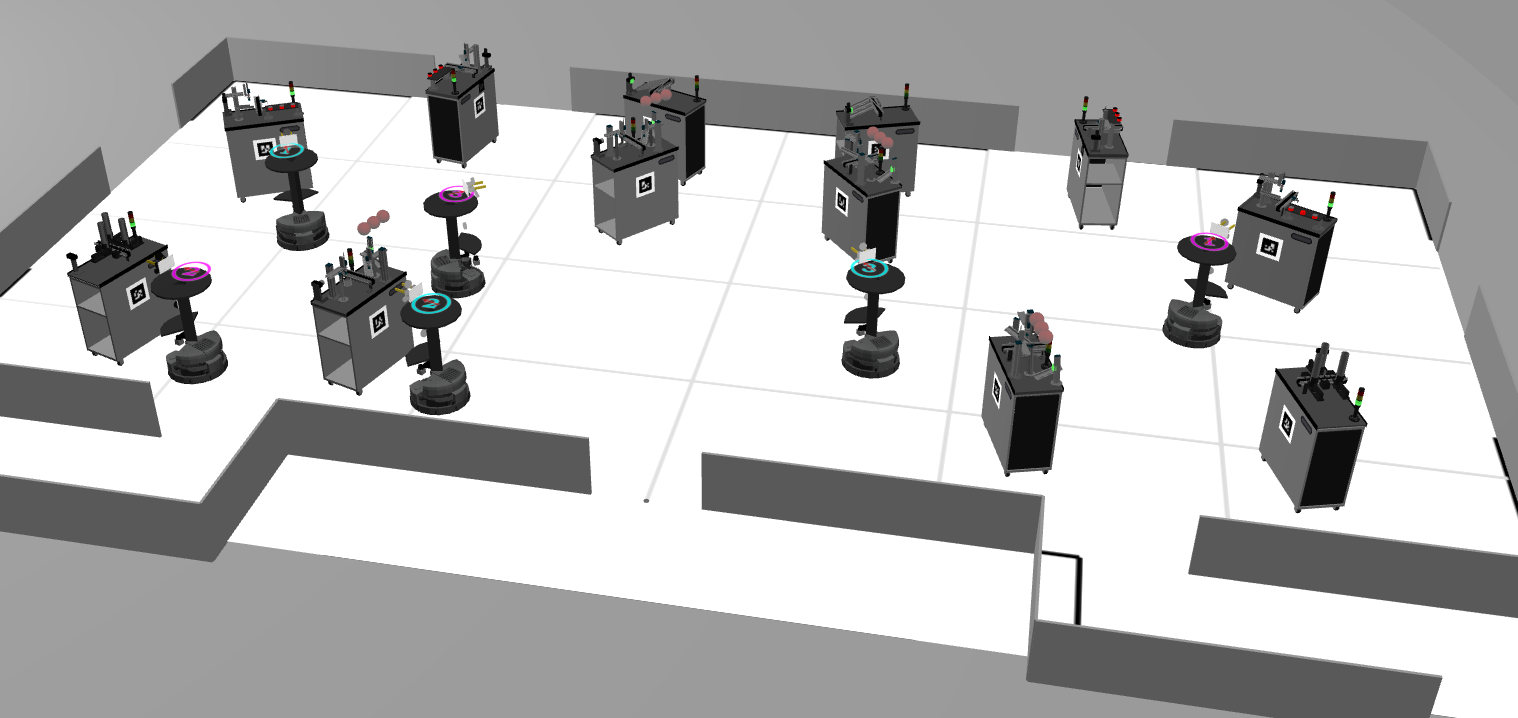}
	\end{center}
	\caption{The simulated RCLL environment.}
	\label{fig:rcll2016-sim}
\end{figure}

We consider a part of the problem posed to the participants of the
\emph{Planning Competition for Logistics Robots in
  Simulation}~\cite{LogRobComp2016} of the RoboCup Logistics League.
The simulated environment is illustrated in Fig.
\ref{fig:rcll2016-sim}. Each game involves two teams, consisting
of three robots each.  The teams do not need to interact with each
other but they execute on the same field.  There are $12$ machines
placed on the field. Each team owns $6$ of the machines on which they
need to operate to produce products according to some
dynamically issued orders.

The field is divided into $24$ zones, $12$ per team. Each team knows that $6$ of its $12$ zones
contain a machine and the other $6$ have no machines in it. 
Before the production can start, each team needs to \emph{explore} its
assigned zones in order to identify the locations and the orientations of their
machines. The exploration must be completed within $20$ minutes.

In this work we focus on the above-described exploration phase. In
this phase, we need to compute a \emph{plan}, which assigns to each of
the three robots of a team a sequence of zones to be visited, such
that each of the $12$ team zones is visited by exactly one of the
robots. For a successful exploration, the computation and the
execution of these plans must be completed within the deadline.
Therefore, the goal is to identify plans with sufficiently short (in best
case \emph{minimal}) total execution time.

The problem formulation is simple, but its solution is
highly challenging (note that the problem is a variation of the multiple traveling
salesman problem). Currently
used solutions are heuristic and cannot assure that the computed solutions are optimal.
Even if such plans allow to complete the exploration before the deadline, optimal plans might
terminate considerably before the deadline and thereby allow to reduce energy
consumption and give the team more time to prepare for the following production phase
of the game.

Our aim is to analyze the applicability of OMT solving
in this context, \ie, to analyze whether optimal solutions can be
determined using OMT solving in a sufficiently short time, such that the computed plan
can still be safely executed in the remaining time before the
exploration deadline.

\section{OMT-based Planning}
\label{sec:encoding}

The experimental analysis that follows has been carried out using
\texttt{Z3}, a well-established OMT solver, on a machine running
Ubuntu Mate 16.4, Intel Core i7 CPU at 2.10GHz and 8GB of RAM.  We
used the \texttt{Python} API of \texttt{Z3} in most cases, with a
single exception of the model validation in Section \ref{sec:optim},
where we used the \texttt{C++} API to integrate our OMT-based solution
into the RCLL framework.

\begin{table*}[]
	\begin{center}
		\begin{tabular}{| l | c | c | c |c | c | c | c | c |c | c | c | c | c |}
			\hline
			\multirow{2}[4]{*}{$\M$} & \multicolumn{2}{ |c| }{$A$} & \multicolumn{2}{ |c| }{$B$} & \multicolumn{2}{ |c| }{$C$} & \multicolumn{2}{ |c| }{$D$} & \multicolumn{2}{ |c| }{$E$} & \multicolumn{2}{ |c| }{$F$} &  \multirow{2}[4]{*}{Optimum}\\
			\cline{2-13} 
			& Time & Conf & Time & Conf & Time & Conf & Time & Conf & Time & Conf & Time & Conf &  \\
			\hline
			6 & 0.40 & 4841 & 0.25 & 3206 & 0.18 & 2525 & 0.17 & 2069 & 0.29 & 3416 & 0.16 & 1103 &10.9   \\
			\hline
			8 & 2.07 & 14400 & 1.91 & 15248 & 1.16 & 9237 & 1.62 & 14355 & 5.32 & 30302 & 1.23 & 3876 & 11.4 \\
			\hline
			10 & 80.06 & 225518 & 59.71 & 184685 & 26.71 & 91648 & 21.72 & 89785 & \multicolumn{2}{c|}{TO} & 8.97 & 27811 & 12.1 \\
			\hline 
			12  & 286.70 & 486988 & 255.55 & 449485 & 81.64 & 198249 & 54.17 & 161134 & \multicolumn{2}{c|}{TO} & 36.21 & 101308 & 12.6\\
			\hline
		\end{tabular}
		
		\caption{Running times (sec) and \#conflicts for encodings A-F ($\M$: number of zones to be visited, TO: 5min). }
		\label{tab:encodings}
	\end{center}
\end{table*}

\medskip

\noindent\textbf{A: First encoding.}\quad
We encode the planning task from Section~\ref{sec:robots} to explore
$\M$ zones by $3$ robots. Robots start from a depot, modeled by some
\emph{fictitious zones} $-3,-2,-1$. Each robot $i\in\{1,2,3\}$ starts at zone
${-}i$, moves over to the zones ${-}i{+}1,\ldots,0$, and explores, from
the \emph{start zone} $0$, at most $\M$ of the zones $1,\ldots,\M$. The distance between two zones $i$ and $j$ is denoted by $\dist(i,j)$. Here we assume the distance that a robot needs to travel to reach the start zone to be $0$, but it could be also set to any positive value (see Fig. \ref{fig:init}). 

The
movements of robot $i$ are encoded by a sequence
$\pos{i}{-i},\ldots,\pos{i}{\M}$ of zones it should visit, with
$\pos{i}{j} \in \mathbb{Z}$. The variables $\pos{i}{-i},\ldots,\pos{i}{0}$
represent the movements from the depot to the start zone:

\begin{figure}[t]
	\begin{center}
		\scalebox{0.9}{\begin{tikzpicture}[->,>=stealth',shorten >=1pt,auto,node distance=3.5cm,
semithick]
\tikzstyle{every state}=[draw=none,text=black,node distance=2cm]

\node[state,fill=green!10]         (A) {$0$};
\node[state,fill=red!10] 		   (B) [below of=A, node distance=2cm]  {$-1$};
\node[state,fill=red!10]         (C) [ right of=B] {$-2$};
\node[state,fill=red!10]         (D) [ right of=C] {$-3$};
\node[state,fill=blue!10]         (E) [above left of=A, xshift=-3ex]  {$1$};
\node[state,fill=blue!10]         (F) [right of=E, node distance=4cm]       {$M$};
\node        	 		(dots) [right of=E, node distance=2cm]       {{$\bm{\ldots$}}};

\path (D) edge              node[rotate=45, yshift=0.55cm,xshift=1cm] {\small$\dist(-3,-2) = 0$} (C)
(C) edge              node[rotate=45, yshift=0.55cm,xshift=1cm] {\small$\dist(-2,-1) = 0$} (B)
(B) edge              node {\small$\dist(-1,0) = 0$} (A)
(A) edge              node[right, yshift=-0.3cm] {\small$\dist(0,M) \neq 0$} (F)
(A) edge              node[left, yshift=-0.3cm] {\small$\dist(0,1) \neq 0$} (E);
\end{tikzpicture}}
	\end{center}

	\caption{Initial robot configuration.}
	\label{fig:init}
	
\end{figure}
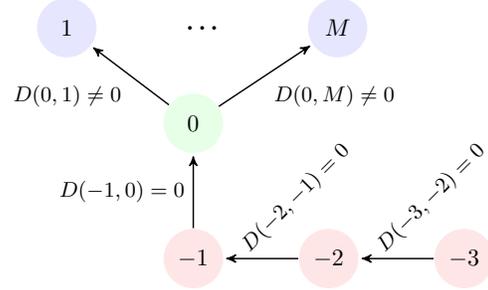

\vspace*{-2ex}
{\footnotesize
\begin{eqnarray}
\label{eq:init_pos}
\begin{array}{l@{}l@{\hspace*{0.5ex}}l@{\hspace*{0.5ex}}l@{}l@{\hspace*{0.5ex}}l@{\hspace*{0.5ex}}l@{}l@{\hspace*{0.5ex}}l@{\hspace*{0.5ex}}l@{}l@{\hspace*{0.5ex}}l}
&&&&&
&\pos{1}{-1} & = -1& \wedge 
&\pos{1}{0}  & =  0&\wedge\\[0.5ex]
&&
&\pos{2}{-2} & = -2&\wedge
&\pos{2}{-1} & = -1&\wedge
&\pos{2}{0}  & =  0&\wedge\\[0.5ex]
 \pos{3}{-3} & = -3&\wedge
&\pos{3}{-2} & = -2&\wedge
&\pos{3}{-1} & = -1&\wedge
&\pos{3}{0}  & =  0&\\
\end{array}
\end{eqnarray}
}

\vspace*{-1ex}
\noindent For $j>0$, if the value of $\pos{i}{j}$ is between $1$ and $\M$ then it encodes the $j$th zone visited by robot $i$. Otherwise, $\pos{i}{j} = -4$ encodes that the robot stopped moving and stays at $\pos{i}{j-1}$ for the rest of the exploration (\ie, the plan does not require robot $i$ to explore any more zones). The total distance traveled by
robot $i$ to visit zones until step $j$ is stored in $\dpart{i}{j} \in \mathbb{R}$. These facts are encoded for each robot $i\in\{1,2,3\}$ by $\dpart{i}{0}=0$ and for each $j\in \{1,\ldots,\M\}$ by:

\vspace*{-2ex}
{\footnotesize
\begin{eqnarray}
\label{eq:move}
\bigvee_{k=0}^{\M} \bigvee_{\substack{l=1\\ l \neq k}}^{\M} && \bigg( \pos{i}{j-1}  {=} k \wedge \pos{i}{j} {=} l \wedge \dpart{i}{j} {=} \dpart{i}{j-1} {+} \dist(k,l) \bigg)\vee\nonumber
\\[-3ex] &&   \bigg( \pos{i}{j}   {=} -4 \wedge \dpart{i}{\M} {=} \dpart{i}{j-1} \bigg)
\end{eqnarray}
}

\vspace*{-2ex}
\noindent which ensures that, at each step $j$, either the robot moves and its travel distance is incremented accordingly, or the robot stops moving. Notice that in this second case, we can immediately determine the final travel distance for the robot at the last step in the plan and, furthermore, the above constraints imply that once a robot stops moving it will not move in the future ($\pos{i}{j}   {=} -4$ implies $\pos{i}{j'}   {=} -4$ for all $j'>j$).

For each zone
$k\in\{1,\ldots,\M\}$ we enforce that it is visited
exactly once by requiring:

\vspace*{-1ex}
{\footnotesize
\begin{eqnarray}
\bigvee_{i=1}^{3} \bigvee_{j=1}^{\M} \bigg( \pos{i}{j} = k  \wedge \bigwedge_{u=1}^{3} \bigwedge_{\substack{v=1 \\ (v,u) \neq (i,j)}}^{\M} \pos{u}{v} \neq k\bigg)
\end{eqnarray}
}

\vspace*{-1ex}
\noindent Finally, for each robot $i\in\{1,2,3\}$ we introduce the Boolean variable $\m{i}$
to encode whether the total travel distance $\dpart{i}{\M}$ for the robot at the end of its plan  is the maximum over
all total travel distances:

\vspace*{-2ex}
{\footnotesize
	\begin{eqnarray}
	\m{i} \Leftrightarrow \bigg(\bigwedge_{\substack{l = 1 \\ l< i}}^{3} \dpart{l}{\M} < \dpart{i}{\M} \wedge \bigwedge_{\substack{l = 1 \\ i<l}}^{3} \dpart{l}{\M} \leq \dpart{i}{\M} \bigg)
	\end{eqnarray}
}

\vspace*{-1ex}
\noindent Our optimization objective is to minimize the largest total travel distance: 

\vspace*{-3ex}
{\footnotesize
	\begin{eqnarray}
	\label{eq:obj2}
	\text{minimize} \quad \sum_{i = 1}^{3} \m{i} \cdot \dpart{i}{\M}
	\end{eqnarray}
}

\noindent\textbf{Results.}\quad We consider four benchmarks with $6$, $8$, $10$ and $12$ zones to be visited. Encoding A allowed us to compute optimal plans, but it does not scale with the number of zones to be visited. The solving time $286.7$ seconds listed in Table~\ref{tab:encodings} for the optimal objective $12.6$ for a benchmark with $\M=12$ zones claims a large part of the overall duration of the exploration phase. \\

\smallskip
\noindent\textbf{B: Tackling loosely connected constraints.}\quad By analyzing solver statistics we noticed that the number of theory conflicts was quite large, and theory conflicts typically appeared at relatively high decision levels, \ie, at late stages of the Boolean search in the SAT solver. 
One reason for this is that during optimization, violations of upper bounds on the total travel distances can be recognized by the theory solver only if all the zones that a robot should visit are already decided. In other words, the constraints
defining the total travel distance of a robot build a loosely connected chain in their variable-dependency graph. Furthermore, explanations of the theory conflicts blamed the whole plan of a robot, instead of restricting it to prefixes that already lead to violation. As a result, the propositional search tree could not be efficiently pruned. To
alleviate this problem, we added to the encoding A for all $i \in \{1,2,3\}$ the following formula, which is implied by the monotone increment of the partial travel distances by further zone visits:

\vspace*{-2ex}
{\footnotesize
\begin{eqnarray}
	\bigwedge_{j = 1}^{\M} \dpart{i}{j} \leq \dpart{i}{\M} \quad 
\end{eqnarray}
}

\vspace*{-1ex}
\noindent\textbf{Results.}\quad As Table~\ref{tab:encodings} shows, adding the above constraints led to a slight improvement, but the solving time of $255.55$ seconds for $12$ zones is still too long for our application.\\

 \begin{figure*}
 	\begin{center}
 		\subfloat[]
 		{
 			\includegraphics[width=0.43\textwidth]{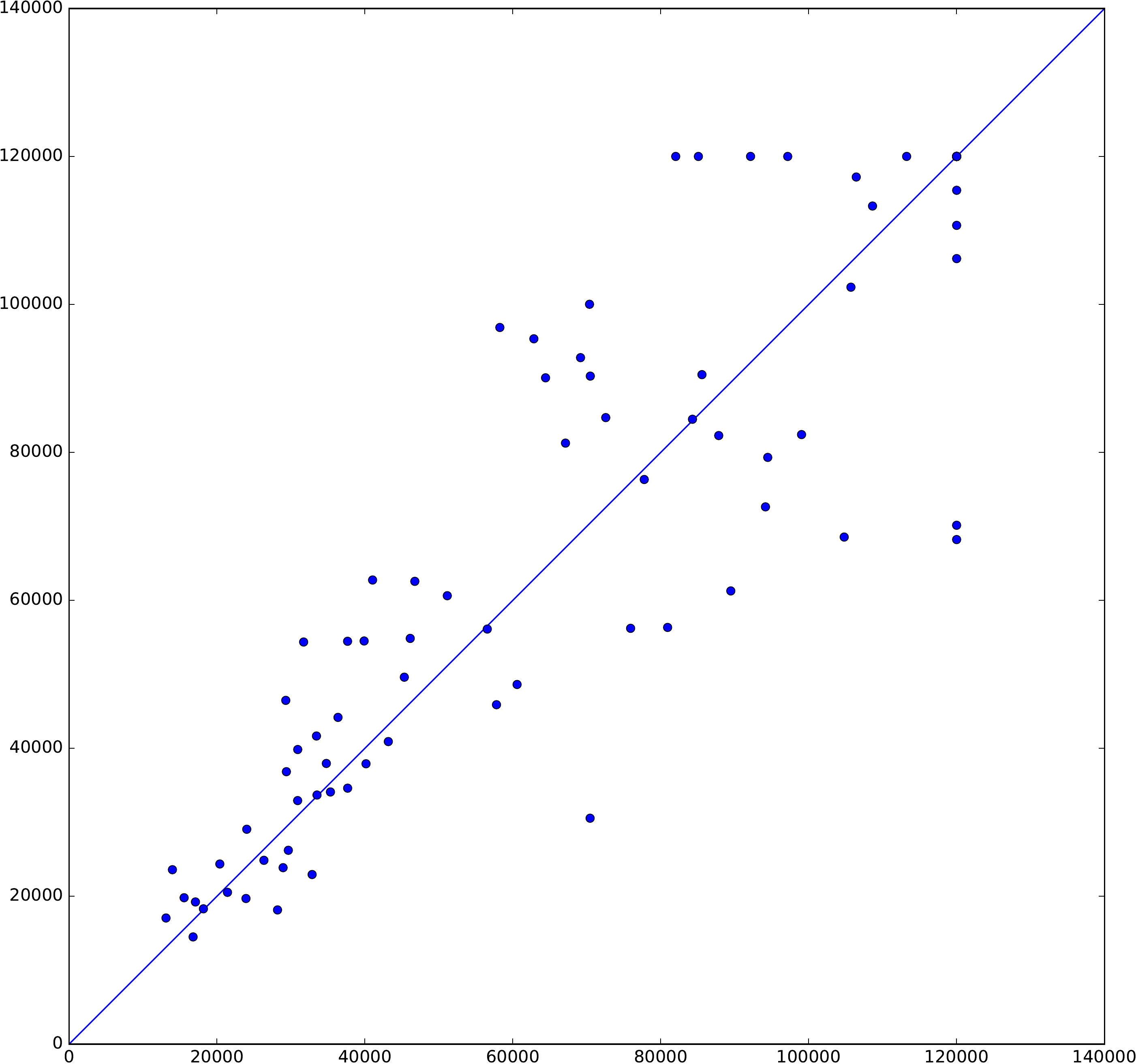}
 			\put(-120,-10){\footnotesize Solving time F}
 			\put(-225,70){\rotatebox{90}{\footnotesize Solving time F$_1$}}
 			\label{fig:a}
 		}
 		\hspace{0.3cm}
 		\subfloat[]
 		{
 			\includegraphics[width=0.43\textwidth]{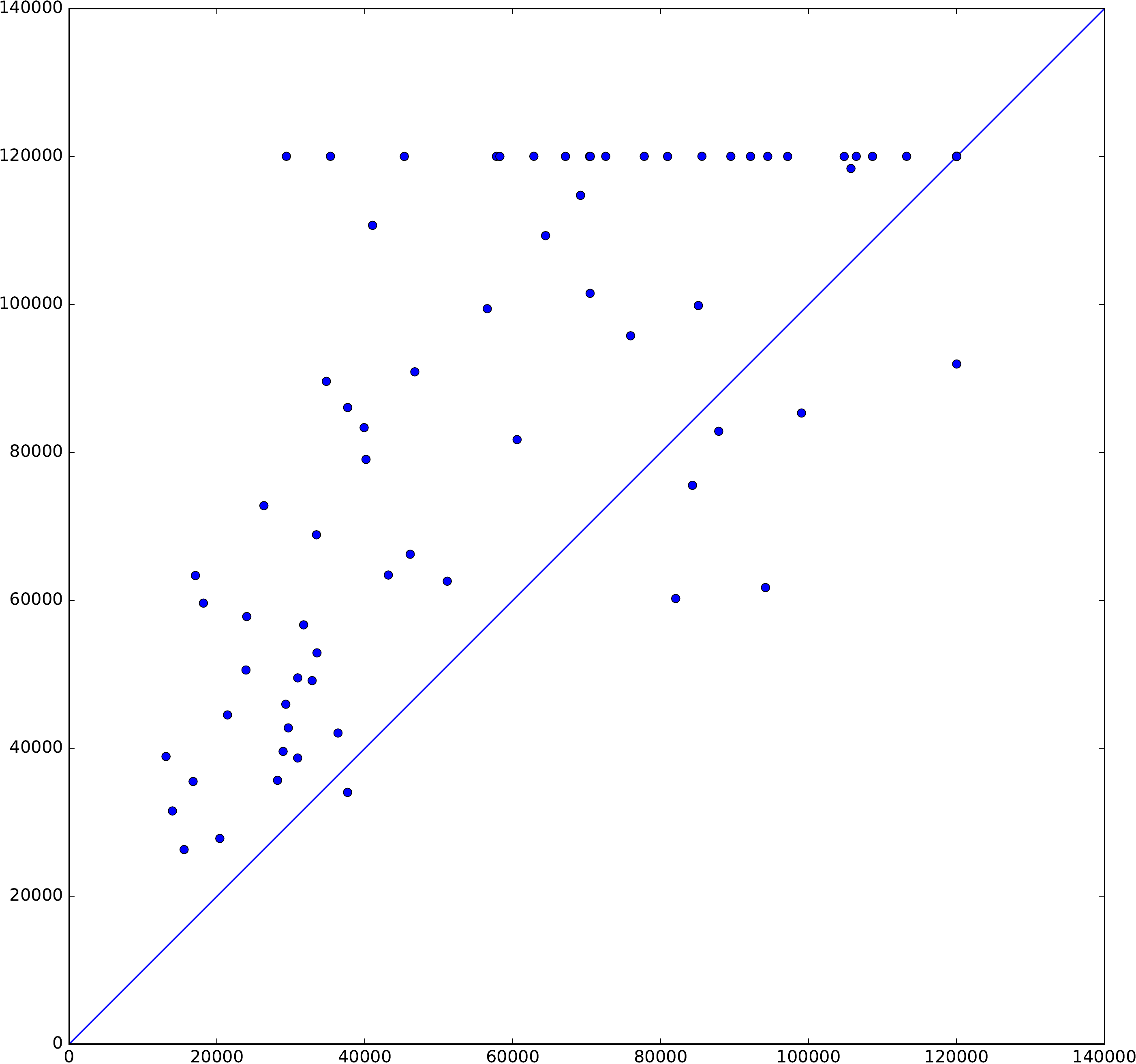}
 			\put(-120,-10){\footnotesize Solving time F}
 			\put(-225,70){\rotatebox{90}{\footnotesize Solving time F$_2$}}
 			\label{fig:b}
 		}
 	\end{center}
 	\caption{Comparison of solving times (msec) for encodings F, F$_1$ and F$_2$ (TO: 2min).}
 	\label{fig:stats}
 \end{figure*}

\smallskip
\noindent\textbf{C: Symmetry breaking.} \quad Although the robots start from different zones, all move to the start location $0$ at cost $0$ before exploration. Thus, given a schedule for the three robots, a renaming of the robots gives another schedule with the same maximal travel distance. These symmetries result in the solver covering unnecessarily redundant search space, significantly increasing solving time. However, breaking these symmetries by modifying the encoding and without modifying the solver-internal algorithms is hard. A tiny part of these symmetries, however, can be broken by imposing on top of encoding B that a single, heuristically determined zone $k$ (\eg, the closest or furthest to zone $0$) should be visited by a robot $i$:

\vspace*{-2ex}
{\footnotesize
\begin{eqnarray}
	\bigvee_{j = 1}^{\M} \pos{i}{j} = k
\end{eqnarray}
}

\vspace*{-1ex}
\noindent\textbf{Results.}\quad This at first sight rather weak symmetry-breaking formula proved to be beneficial, resulting in a greatly reduced number of conflicts as well as solving time ($81.64$ seconds for $\M=12$ zones, see Table \ref{tab:encodings}). However, this encoding just fixes the robot that should visit a given single zone, thus the computational effort for $\M$ zones reduces only to a value comparable to the previous effort (using encoding B) for $\M-1$.\\

\smallskip
\noindent\textbf{D: Explicit scheduler choice.}\quad
In order to make the domain over which the variables $\pos{i}{j}$ range more explicit, we added to encoding C the following constraints for all $i\in\{1,2,3\}$ and $j \in\{1,\ldots,\M\}$ :

\vspace*{-2ex}
{\footnotesize
	\begin{eqnarray}
	\pos{i}{j}=-4\vee \bigvee_{k = 1}^{\M} \pos{i}{j} = k 
	\end{eqnarray}	
}

\vspace*{-1ex}
\noindent\textbf{Results.}\quad This addition led to some performance gain. With a solving time of $54.17$ seconds for $12$ zones, our approach could be successfully integrated in the RCCL planning framework.\\

\smallskip
\noindent\textbf{E: Partial bit-blasting.}\quad To reduce the number and size of theory checks, we also experimented with partial bit-blasting: the theory constraints $\pos{i}{j}{=}k$ in encoding C were replaced by Boolean propositions $\posb{i}{j}{k} {\in} \mathbb{B}$, which are true iff robot $i$ visits zone $k$ at step $j$. For each $i{\in}\{1,2,3\}$ and  $j{\in}\{-3,{\ldots},\M\}$ we ensure that there is exactly one $k{\in}\{-4,{\ldots},\M\}$ for which $\posb{i}{j}{k}$ is true by bit-blasting for the $\M{+}5$ possible values (using fresh propositions ${p_{i,j,k} \in \mathbb{B}}$):

\vspace*{-2ex}
{\footnotesize
\begin{eqnarray}
\posb{i}{j}{0} &\iff& (\neg p_{i,j,\ceil{\log{(\M+5)}}} \wedge \ldots \wedge \neg p_{i,j,0})\nonumber\\
\posb{i}{j}{1} &\iff& (\neg p_{i,j,\ceil{\log{(\M+5)}}} \wedge \ldots \wedge \phantom{\neg} p_{i,j,1})\quad\ldots
\end{eqnarray}
}

\vspace*{-2ex}
\noindent\textbf{Results.}\quad As shown in Table~\ref{tab:encodings}, partial bit-blasting did not introduce any improvement. On the contrary, an optimal solution for $12$ zones could not be computed within $5$ minutes. We made several other attempts to improve the running times by modifying encoding $D$, but they did not bring any major improvement.\\

\smallskip
\noindent\textbf{F: Explicit decisions.} Even though encoding D could be integrated in the RCLL framework, we investigated ways to further reduce the solving times. 

To this purpose, we developed a new encoding, in which 
we made some decisions explicit by means of additional variables. In particular, we introduced integer variables $\m{k}$ to encode which robot visits zone $k$, and integer variables $\nm{i}{k}$ to count how many of the zones $1, \ldots, k$
 robot $i$ has to visit.
For each $i \in \{1, 2, 3\}$, we initialize $\nm{i}{0} = 0$ and enforce for each $k \in \{1, \ldots , \M\}$:

\vspace*{-2ex}
{\footnotesize
	\begin{eqnarray}
	(\m{k} = i \wedge \nm{i}{k} = \nm{i}{k-1} + 1) \vee (\m{k} \neq i \wedge \nm{i}{k} = \nm{i}{k-1})
	\end{eqnarray}
}

\vspace*{-2ex}
\noindent We keep the position variables $\pos{i}{j}$ to store which zone is visited in step $j$ of robot $i$, but their domain is slightly modified:
knowing the number $\nm{i}{\M}$ of visits for each robot, the fictitious location $\pos{i}{j} = -4$ is not needed anymore. Instead, we will simply disregard all $\pos{i}{j}$ assigned for ${j > \nm{i}{\M}}$. 

We also keep the variables $\dpart{i}{j}$, but with a different meaning: $\dpart{i}{j}$ stores the distance traveled by robot $i$ from its $(j{-}1)$th position $\pos{i}{j-1}$ to its $j$th position $\pos{i}{j}$. We add the constraints (\ref{eq:init_pos}) for defining the positions up to the start zone and the following constraints for $i \in \{1, 2, 3\}$ and $j \in \{1, \ldots , \M\}$:

\vspace*{-2ex}
{\footnotesize
	\begin{eqnarray}
	\label{eq:new_move}
	\bigvee_{k=0}^{\M} \bigvee_{\substack{l=1\\ l \neq k}}^{\M} \bigg( \pos{i}{j-1}  {=} k \wedge \pos{i}{j} {=} l \wedge \dpart{i}{j} \geq \dist(k,l) \bigg)
	\end{eqnarray}
}

\vspace*{-1ex}
\noindent 
Note that instead of $\dpart{i}{j} = \dist(k,l)$ we added inequalities because this improved the solving times, while minimization will anyways enforce equality (see also page \pageref{sec:ineq}). 

A new variable $\dtot{i}$ for $i \in \{1,2,3\}$ is used to store the total travel distance for each robot.  

We enforce for $i \in \{1,2,3\}$:

\vspace*{-2ex}
{\footnotesize
	\begin{eqnarray}
	\label{eq:bound}
	(\nm{i}{\M} = 0 \wedge \dtot{i} \geq 0) \vee \bigvee_{k = 1}^{\M} (\nm{i}{\M} = k \wedge \dtot{i} \geq \sum_{j = 1}^{k} \dpart{i}{l})
	\end{eqnarray}
}

\vspace*{-1ex}
\noindent which ensures that, if robot $i$ has to visit $k$  zones ($\nm{i}{\M} = k$) then its total travel distance $\dtot{i}$ is (at least equal to) the sum of the distances traveled from $\pos{i}{0}$ to $\pos{i}{k}$. If robot $i$ does not move at all (\ie, $\nm{i}{\M} = 0$) then $\dtot{i}$ will be (at least) zero.

In order to make sure that each robot visits all zones it has been assigned to (by means of variables $\m{k}$), we add the following constraint for each $i \in \{1,2,3\}$ and $k \in \{1, \ldots, \M\}$:

\vspace*{-2ex}
{\footnotesize
	\begin{eqnarray}
	\m{k} = i \implies \bigvee_{j = 1}^{\M} \nm{i}{\M} \geq j \wedge \pos{i}{j} = k
	\end{eqnarray}
}

\vspace*{-1ex}
\noindent which ensures that, if robot $i$ is assigned to zone $k$ then this zone will be visited at some step $j$ (within the upper bound on the number of zones to be visited $\nm{i}{\M}$).

Furthermore, we introduce bounds on integer variables so that the solver can internally perform bit-blasting and represent them as bit-vectors. As a consequence, the following constraints were added for each $i \in \{1,2,3\}$ and ${k,j \in \{1, \ldots, \M\}}$:

\vspace*{-2ex}
{\footnotesize
	\begin{eqnarray}
	\label{eq:int_bound}
	1 \leq \m{k} \leq 3 \qquad 0 \leq \nm{i}{k} \leq \M \qquad 1 \leq \pos{i}{j} \leq \M
	\end{eqnarray}
}

\vspace*{-1ex}
Finally, we replace the nonlinear objective function specified in constraint (\ref{eq:obj2}) by a linear one: since all robots start from the start zone, we exploit symmetry and require an order on the total travel distances:

\vspace*{-2ex}
{\footnotesize
	\begin{eqnarray}
	\label{eq:symmm}
	\dtot{1} \geq \dtot{2} \wedge \dtot{2} \geq \dtot{3}
	\end{eqnarray}
}  

\vspace*{-1ex}
\noindent We can now minimize the total distance for the first robot:

\vspace*{-2ex}
{\small
	\begin{eqnarray}
	\label{eq:new_obj}
	\text{minimize } \dtot{1} 
	\end{eqnarray}
}  

\vspace*{-1ex}
\noindent\textbf{Results.} \quad Table \ref{tab:encodings} shows a considerable improvement by encoding F over previous solutions for the selected benchmarks. 
In order to obtain statistically significant results, we also tested encoding F on $100$ most recurring instances of the RCLL problem with $12$ zones (see Table \ref{tab:encodings_2}). Three optimizing solvers were used to carry out this series of experiments, namely \texttt{Z3}, \texttt{SMT-RAT} and \texttt{OptiMathSAT}. The latter specializes in optimization for real arithmetic while \texttt{SMT-RAT} is tuned for non-linear real arithmetic problems. However, our problem mostly involves optimization at the Boolean level and therefore the strengths of these two solvers could not be exploited to their fullest. For this reason, we report only results obtained using \texttt{Z3}.

\label{sec:ineq}
Replacing a non-linear objective function with a linear one allowed us to reduce the complexity of the optimization problem at hand. Moreover, specifying bounds on integer variables in (\ref{eq:int_bound}) and specifying lower bounds instead of equalities in (\ref{eq:new_move}) and  (\ref{eq:bound}) seemed to be also helpful.
To analyze these potential sources of improvement, we made additional experiments with two variants of encoding $F$: in encoding F$_1$ we removed the bounds (\ref{eq:int_bound}) for integer variables, and in encoding F$_2$ we replaced the inequalities in (\ref{eq:new_move}) and (\ref{eq:bound})  by equalities (but the constraints (\ref{eq:int_bound}) are kept in F$_2$). Table \ref{tab:encodings_2} shows the average results for the previously used $100$ benchmarks.
While working with unbounded integers does not seem to significantly affect the solving times (Fig.~\ref{fig:a}), relaxing equalities had a stronger impact. Fig.~\ref{fig:b} shows that the solving time for the encoding F$_2$ with equalities is almost always higher, and a fewer number of instances could be solved within the timeout (see Table~\ref{tab:encodings_2}).

\begin{table}[t]
	\begin{center}
		\begin{tabular}{| l | c | c | c | c| c | c | }
			\hline
			\multirow{2}[4]{*}{$\M$} & \multicolumn{2}{ |c| }{$F$} & \multicolumn{2}{ |c| }{F$_1$} & \multicolumn{2}{ |c| }{F$_2$} \\
			\cline{2-7} 
			& Time & \#solved & Time & \#solved & Time & \#solved  \\
			\hline 
			12  & 54.78 & 66/100 & 57.02 &  66/100 & 66.84 & 46/100  \\
			\hline
		\end{tabular}
		\caption{Average solving time (sec) and  \#instances solved for encodings F, F$_1$  and F$_2$ on $100$ benchmarks (TO: 2min).}
		\label{tab:encodings_2}
	\end{center}
\end{table}

\section{Additional Notes}
\label{sec:optim}
 
\noindent\textbf{About the optimization algorithm.}\quad \texttt{Z3}
adds the clauses shown in Eq.~(\ref{eq:optimization}) to $(i)$ exclude
the current assignment $S$ from the further search and $(ii)$ improve
the current upper-bound value
$\mu$~\cite{DBLP:conf/tacas/BjornerPF15}. However, as already noted
in~\cite{DBLP:conf/cade/SebastianiT12,DBLP:journals/tocl/SebastianiT15},
the inconsistency of the arithmetic constraint $f < \mu$
cannot be determined at the Boolean level. Therefore,
inconsistencies on this term are not detected until the theory solver
is invoked. Since the latter is more resource demanding and invoked
less often, the performance of a solver based on
linear-search optimization can be strongly affected when
dealing with a complex and unstructured domain like the one at hand.

To examine these aspects in more detail, we implemented two basic optimization algorithms \texttt{Simple} and \texttt{Binary} with three different versions \texttt{Opt1}, \texttt{Opt2} and \texttt{Opt3} each, based on the satisfiability checking function of \texttt{Z3} (instead of its optimization function). For all versions of both algorithms, the input formula $\varphi$ is constructed as in encoding C but without the optimization constraint (\ref{eq:obj2}). 

For each found solution $S$ with value $\nu$ for the objective function $f$, \texttt{Simple} adds a formula, similar to Equation~(\ref{eq:optimization}) and described below, to reduce the further search to values from $[0,\nu)$ for $f$; note that for Equation~(\ref{eq:optimization}) we would compute the optimal value $\mu$ under $S$ and add $f<\mu$, however, for our problems $\nu$ will always be a local optimum as its value is defined by equations in encoding C. \texttt{Simple} terminates with the last found solution when the refined formula becomes unsatisfiable.

\texttt{Binary} determines first a solution with value $\nu$ for $f$ and maintains in its further search lower and upper bounds $l$ and $u$ on the optimal value, which are initially $0$ and $\nu$, resp. In each further iteration, \texttt{Binary} first searches for a solution with an objective function value $\nu'$ in the lower half $[l,m)$ of the interval $[l,u)$. If a solution is found then $u$ is set to $\nu'$. Otherwise if there is a solution in the upper half $[m,u)$ with objective function value $\nu'$ then $l{:=}m$ and $u{:=}\nu'$ are set; if neither $[l,m)$ nor $[m,u)$ contain a solution then \texttt{Binary} terminates with the last solution. 

For both algorithms, their three versions differ in the formula that is added for each found objective function value $\nu$ to restrict the further search. For \texttt{Simple}, version \texttt{Opt1} extends the formula $\varphi$ by adding the constraint $f<\nu$. \texttt{Opt2} adds the formula $f<\nu\wedge\neg \varphi_{\textit{max}}$, where $\varphi_{\textit{max}}$ encodes the sequence of visited zones for the robot with the longest travel distance. Finally, \texttt{Opt3} adds $f<\nu\wedge\neg \varphi_{\textit{max}}\wedge\neg \varphi_{\textit{max}}'\wedge\neg \varphi_{\textit{max}}''$, where $\varphi_{\textit{max}}'$ and $\varphi_{\textit{max}}''$ are copies of $\varphi_{\textit{max}}$ with the index of the robot having the largest travel distance substituted by the indices of the other two robots. That means, \texttt{Opt3} adds further symmetry breaking constraints. For \texttt{Binary}, all three versions are analogous to what was previously described but additionally add the corresponding lower bounds on $f$.\\

\begin{table}
	\begin{center}
		
		\begin{tabular}{| l | c | c | c |}
			\cline{2-4}
			\multicolumn{1}{ l |}{} & \texttt{Opt1} & \texttt{Opt2} & \texttt{Opt3}\\
			\hline
			\texttt{Simple} & 107.10 & 157.85 & 166.57\\
			\hline
			\texttt{Binary} & 117.40 & 104.69 & 95.55\\
			\hline
		\end{tabular}
		\caption{Optimization times (sec) for $\M{=}12$ zones using encoding $C$.}
		\label{tab:opt}
		
	\end{center}
\end{table}
\noindent\textbf{Results.}\quad Table \ref{tab:opt} shows results for the $12$ zones benchmark used for Table \ref{tab:encodings}. The optimization algorithm in \texttt{Z3} is (unsurprisingly) more efficient ($81.64$ sec, see encoding C for $\M=12$ in Table \ref{tab:encodings}) than our prototypical algorithms. Nevertheless, these algorithms allow us to compare different optimization alternatives. Especially, the results show that excluding the current solution on top of restricting the domain for the objective function value pays off only for the binary search \texttt{Binary}, but not for \texttt{Simple}.\\
%
%
%
%

\noindent\textbf{Comparison to other planning methods.}\quad The results presented so far are encouraging as they show potential to increase scalability. However, the improved solving times are still high compared to other techniques. To illustrate this, we compare our results to traditional planning with PDDL, an action-centered language, inspired by \texttt{STRIPS} formulations of planning (see \cite{DBLP:journals/jair/FoxL03} for a complete description). PDDL planning, being heuristic-driven, is very fast but in general does not compute optimal solutions. For example, for the benchmark used in Table~\ref{tab:encodings} for $\M=12$ zones, the \texttt{POPF} planner~\cite{DBLP:conf/aips/ColesCFL10} returns a schedule with objective function value $20.54$ in $0.04$ seconds, in contrast to the optimum $12.6$ determined by OMT solving.

OMT-based methods are in general computationally more expensive than currently used planning techniques. Nevertheless, the optima found by OMT-based methods are far better than the solutions found by other planning techniques. The latter scale well in terms of efficiency, but give no guarantees on the quality of the computed solutions. We therefore investigated the effects of combining the two techniques. Given an instance of the RCLL, \texttt{POPF} is used first to compute an upper bound $d_{max}$ on the maximum  distance traveled by robots (as per Eq.~(\ref{eq:new_obj})). Then, an SMT-LIB~\cite{BarFT-SMTLIB} encoding is created for the same instance and \texttt{Z3} is called with the added constraint $d_1\leq d_{max}$.\\

\noindent\textbf{Results.}\quad Table~\ref{tab:comp_planner} shows the effect of introducing the upper bound computed using planning. Results refer to encoding F without and with upper bound -- the latter being called F$_{UB}$. At a first glance, it may look reasonable to conclude that having upper bounds on objective functions is beneficial: the solver could solve more instances and with an average time of $50.13$ seconds, which is less than what was previously obtained. We therefore carried out Mann-Whitney U test on the $66$ instances which were already solvable with encoding F to verify whether the above statement is statistically relevant. 
Mann-Whitney U test allows to test a null hypothesis that it is equally likely that a randomly selected value from one sample (solving times for F) will be less than or greater than a randomly selected value from a second sample (solving times for F$_{UB}$). 
Running this test, a \textit{p}-value of $0.18$ was obtained, which does not allow us to discard the null hypothesis within a reasonable significance level (\eg, $5\%$).

Further inspection of the results revealed that $27\%$ of the instances which were already solvable with encoding F see an increase in solving time, with an average rising from $66.75$ seconds to $78.87$ seconds. We conjecture that reducing the search space over the reals does not necessarily lead to an improvement in solving time, as the number of feasible computation traces reduces therefore making the problem harder to solve. This point will be the subject of future investigations.\\

\begin{table}
	\begin{center}
		\begin{tabular}{| l | c | c | c | c|  }
			\hline
			\multirow{2}[4]{*}{$\M$} & \multicolumn{2}{ |c| }{$F$} & \multicolumn{2}{ |c| }{$F_{UB} $}  \\
			\cline{2-5} 
			& Time & \#solved & Time & \#solved   \\
			\hline 
			12  & 54.78 &  66/100 & 50.13 & 70/100  \\
			\hline
		\end{tabular}
		\caption{Average solving time (sec) and  \#instances solved for encodings F and F$_{UB}$ (TO: 2min).}
		\label{tab:comp_planner}
	\end{center}
\end{table}

%

\noindent\textbf{Model validation.}\quad In order to assess the feasibility and quality of the plans computed, we executed them in the RCLL simulation environment. We measured the time between start and end of the simulation (solving time excluded) for each plan; given that robots move at a speed between $0.1$ and $0.5$ (distance unit per time unit in simulation),
a conservative estimate of the distance traveled can be obtained by multiplying the above mentioned time by, \eg, $0.1$. The ratio between this estimate and the solution computed by the solver resulted to be $1.42\pm0.72$. This difference is to be expected as planning assumptions are most often challenged during execution, due to unforeseen events and environment uncertainty -- \eg, when two robots get stuck or have to give way to one another. Such situations require the executive to adjust its current plan to cope with the unforeseen scenario, subsequently leading to (possibly) considerable variations. However, it should be noted that such variations occur irrespectively of the planning algorithm used.

\section{Concluding Remarks}
\label{sec:conclusion}

In this paper we have shown how SMT solving combined with optimization can be used to synthesize guaranteed-quality plans for robot fleets. Based on a RoboCup Logistics League case study, we highlighted weaknesses and strengths of our approach. 
The way how problems like the one presented are logically encoded has a strong influence on the solving time. In particular, we observed that improvements on optimization were gained by:
\begin{itemize}
	\item applying symmetry breaking as much as possible;
	\item explicitly encoding transitive properties like for the monotonicity of sequences;
	\item introducing bounds on integer-valued variables so that the solver can apply internal bit-blasting (however, adding bounds for real-valued variables does not always lead to improvements);
	\item avoiding non-linearities whenever possible;
	\item replacing equalities by inequalities if the objective function enforces the equality.
\end{itemize}
For our application the above suggestions led to improvements, and we expect similar effects for applications from other areas. Nevertheless, these observations cannot be generalized, because solver performances might be extremely sensitive  to even minor changes in the encoding. 

Concerning the solvers, while there exist efficient solvers for different types of optimization problems like combinatorial optimization or integer programming problems, there seem to be room for improvements for problems where the objective function is a real-arithmetic function but the search is over a finite set of objects, \ie, where the problem seems to involve optimization in the arithmetic domain but at its core it is a purely combinatorial optimization problem.

Moreover, binary-search-based optimization algorithms seem to be very sensitive to the parameters used to drive the search. This point could be exploited, \eg, to develop a portfolio approach, where different solvers, instantiated with different parameters, run in parallel.

Though our OMT-based planning module has been successfully integrated in the RCLL planning framework, several challenges are yet to be addressed in order to improve the performance when dealing with more complex tasks in the production phase. Future research will investigate ways to better exploit domain-specific knowledge, parallelisation strategies for optimization, and possible ways improve OMT by combining it with planning  heuristics.

\section*{Acknowledgments}
The authors would like to thank Igor Bongartz and Lukas Netz for their help
with the implementation and validation of the results presented.

\bibliographystyle{latex8}
\bibliography{literature}

\end{document}